\documentclass[10pt,twocolumn,letterpaper]{article}

\usepackage[pagenumbers]{cvpr} 
\usepackage{multirow}
\usepackage[table]{xcolor}
\usepackage{pifont}
\usepackage{booktabs}  
\usepackage{nicematrix}
\usepackage{amsmath, amssymb}
\usepackage{algorithm}
\usepackage{algpseudocode}
\usepackage[utf8]{inputenc}
\definecolor{lightgray}{rgb}{.91,.91,.91}
\definecolor{rouse}{rgb}{0.981,0.961,0.941}
\definecolor{deepred}{rgb}{0.698,0.133,0.133}
\definecolor{blue}{rgb}{0,0,1}
\definecolor{mygray}{gray}{0.9}
\definecolor{cvprblue}{rgb}{0.21,0.49,0.74}
\usepackage[pagebackref,breaklinks,colorlinks,allcolors=cvprblue]{hyperref}

\def\paperID{6296} 
\def\confName{CVPR}
\def\confYear{2026}

\title{LEEVLA: Seeing What Matters in \textbf{L}atent \textbf{E}nvironment \textbf{E}volution for Vision-Language-Action}

\author{
Qi Lyu$^{1,2,3}$,
Baicheng Liu$^{1,2}$,
Xudong Wang$^{1,2,3}$,
Jiahua Dong$^{4}$,
Lianqing Liu$^{1,2}$,
Zhi Han$^{1,2}$ \\[1em]
$^{1}$State Key Laboratory of Robotics and Intelligent Systems \\
$^{2}$Shenyang Institute of Automation, Chinese Academy of Sciences \\
$^{3}$University of Chinese Academy of Sciences \\
$^{4}$Mohamed bin Zayed University of Artificial Intelligence \\
}

\begin{document}
\maketitle

\begin{abstract}
Vision-language-action (VLA) models aim to map multimodal inputs to robot actions. However, most existing approaches struggle to cover complex dynamic scenarios due to treating all visual tokens uniformly and reasoning with human-selected factors, which lack mechanisms to emphasize task-critical evidence and ignore underlying factors. To address this issue, we propose \textbf{LEEVLA}, a VLA architecture for seeing what matters in \textbf{L}atent \textbf{E}nvironment \textbf{E}volution that explicitly guides the model toward informative regions while preserving the structured evolution of latent world representations. To identify salient and instruction-relevant regions, we introduce drift-guided dynamic prioritization (DGDP), which combines dynamic position prioritization (DPP) with semantic drift guidance (SDG) to guide the VLA agent where to attend during training. On top of this, we introduce structured feature flow generation (SFFG), which models how these prioritized features should evolve in latent space via prototype-to-periphery (P2P) prediction, and a mutual-neighborhood contrastive (MC) loss to maintain topological consistency among neighborhoods. Together, DGDP and SFFG form a task-aware ``where–how” training framework. Extensive experiments on VLA benchmarks show that LEEVLA consistently outperforms prior methods, confirming that explicit task-evidence guidance and structured latent reasoning are both crucial for scalable VLA. Our code is available in the \url{https://github.com/LyuQi127/LEEVLA}.
\end{abstract}    
\section{Introduction}
\label{sec:intro}

\begin{figure}
\centering
    \includegraphics[width=1\linewidth]{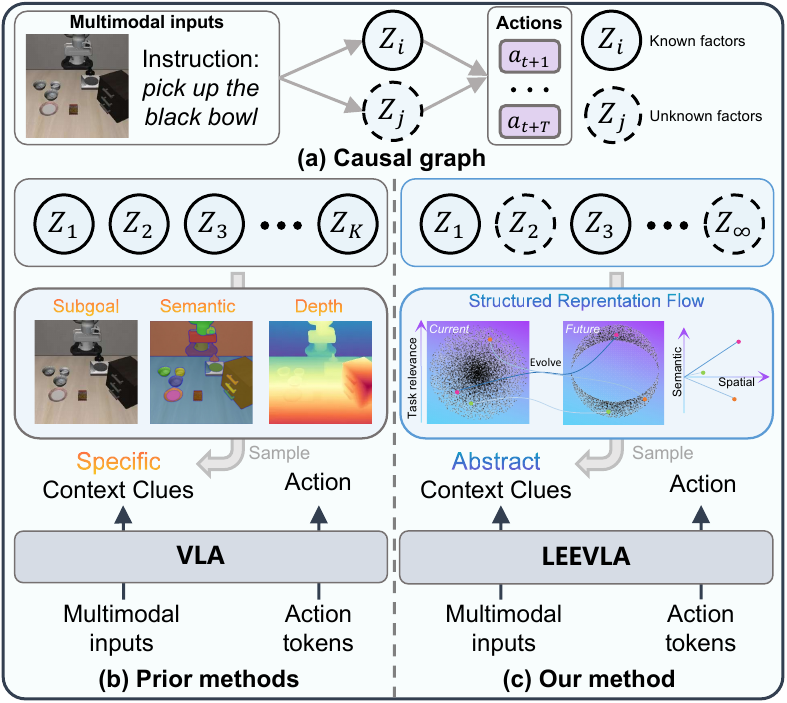}
    \vspace{-5mm}
    \caption{Comparison between our method and prior methods. (a): Causal graph between multimodal inputs and action chunk. (b): Prior methods: VLA reasons on factors of specific context clues. (c): Our method: VLA reasons on abstract context clues.}
    \label{fig:motivation}
    \vspace{-7mm}
\end{figure}

Vision-Language-Action (VLA) \cite{black2025pio, ghosh2024octo, kim24openvla, driess2025knowledgeinsulatingvisionlanguageactionmodels, yang2025instructvlavisionlanguageactioninstructiontuning, zhao2025cotvla, brohan2023rt2, din2025visionlanguageactionmodelssurvey, LBP, hu2025videopredictionpolicygeneralist,SeqWalker} models aim to ground perception and language in action, mapping observation and language instructions directly to low-level controls in the closed loop. By jointly encoding visual observations, proprioception, and language instruction, VLAs offer a path toward robots that can perceive, reason, and act in open environments. Recent research typically employs large language models (LLMs) or visual-language models (VLMs) to construct VLAs \cite{touvron2023llama, openai2022chatgpt, qwen2team2024qwen2,lyu2026sablvlmsignificanceawarebinarizationlarge}, and then trains these VLAs on large-scale robot demonstrations or simulated interaction traces \cite{liu2023liberobenchmarkingknowledgetransfer, openxembodiment2024}. Notably, the generation of action is governed by both known and unknown factors \cite{NEURIPSLearning_Concept}, as illustrated by the causal graph in Fig.~\ref{fig:motivation} (a). Recently, several works have further introduced explicit intermediate reasoning prior to decoding action to strengthen world understanding and task decomposition. As shown in Fig.~\ref{fig:motivation} (b), these methods confine the scope of reasoning to specific context clues selected by humans. Such vision-language reasoning is effective in sharpening spatial perception and making downstream policies more interpretable.
However, those methods based on reasoning still suffer from a restricted search space due to their dependence on external conditions or priors such as subgoal images, segmentation, or depth~\cite{dreamvla25, zhao2025cotvla,TraceVLA}. 

Considering that most available embodied demonstration datasets \cite{liu2023liberobenchmarkingknowledgetransfer, mees2022calvin, zhang2025vlabench, jiang2022vima,james2019rlbench} have restricted modality diversity (e.g., few scene layouts, limited object categories, and limited concepts), methods depending on specific context clues require auxiliary models to produce these modalities. Beyond incurring additional computational overhead, such methods are contingent on progress in auxiliary tasks.
Consequently, these methods~\cite{dreamvla25, zhao2025cotvla} overemphasize known factors while under-exploring unknown but task-relevant factors~\cite{hase-etal-2020-leakage, carton-etal-2022-learn}, leading to fragility when the same object varies across modalities or different objects look similar within a single modality. As a result, such designs can inadvertently limit the exploratory capacity of the model in latent space, making it harder for the backbone to discover task-critical but unknown factors~\cite{wang2025BlindFormer}. 

To address these limitations, we advocate reasoning directly in the latent feature space, treating actions as drivers of how environment states evolve in high-dimensional space, as shown in Fig.~\ref{fig:motivation} (c). The success of general vision models~\cite{zhai2023siglip,Shafiullah2022CLIPFieldsWS,oquab2023dinov2} on all sorts of tasks~\cite{yang2024depthanything, ravi2024sam2, kirillov2023segany} suggests that features produced by general vision models embed multimodal information, such as category and depth. Through reasoning over these latent representations, a policy can jointly exploit semantic, appearance, and geometric cues that are entangled for the same object~\cite{howlearning, Li2024LWE}. Specifically, considering only depth makes it difficult to recover the spatial position of the target object under partial occlusion, while other modalities can help the robot localize it. Operating in latent space also avoids auxiliary pixel-level reconstruction or externally supplied condition pipelines during training, reducing overhead while enlarging search space of the policy to uncover both known and unknown task-relevant factors~\cite{NEURIPSLearning_Concept}. Additionally, naive prediction in a high-dimensional space often diverts attention toward static background or instruction-irrelevant objects, undermining reasoning. Thus, deciding where to attend is equally critical: our insight is to prioritize regions that exhibit pronounced spatial change and whose semantic evolution aligns with the language instruction, so the model concentrates supervision and capacity on scene components that are causally tied to the intended manipulation.

To this end, we propose \textbf{L}atent \textbf{E}nvironment \textbf{E}volution VLA (LEEVLA), in which drift-guided dynamic prioritization (DGDP) tells the agent \textbf{where to attend}, while structured feature flow generation (SFFG) models \textbf{how to evolve} latent environment representation.
LEEVLA adopts drift-guided dynamic prioritization (DGDP) to automatically discover salient and instruction-relevant regions in the feature space via dynamic position prioritization (DPP) and semantic drift guidance (SDG), thereby “focusing” the model on where to attend during training. At the same time, we introduce structured feature flow generation (SFFG) to make the model reason over these regions on how to evolve by enforcing prototype-to-periphery (P2P) prediction. Furthermore, we found that the latent space is inevitably contaminated by spurious neighbors and asymmetric affinity. Similar to the phenomenon observed in passive discriminant analysis, where only reciprocal neighbors or high-affinity neighbors can reliably share semantics, we introduce a mutual-neighborhood contrastive (MC) loss to filter out these noisy links and maintain semantic neighborhood consistency. To sum up, drift-guided dynamic prioritization (DGDP) explicitly steers the model toward task-critical context clues, while structured feature flow generation (SFFG)  preserves the structured evolution of environment representations. Through extensive experimentation, we demonstrate that our approach achieves state-of-the-art performance among existing methods.
Our contributions include the following points:

\begin{itemize}
    \item We propose a drift-guided dynamic prioritization (DGDP) mechanism for automatic discovery of where to attend, composed of dynamic position prioritization (DPP) and semantic drift guidance (SDG), which identifies dynamically active and instruction-relevant regions.
    \item We propose a structured feature flow generation (SFFG) strategy incorporating prototype-to-periphery (P2P) prediction and mutual-neighborhood contrastive (MC) loss, guiding the model to learn how to evolve in latent space. 
    \item Extensive experiments shows consistent performance gains, validating that explicit task-relevance guidance and structured latent reasoning together enhance the generalization and long-horizon reasoning capability of VLA.
\end{itemize}

\begin{figure*}[t]
\centering
    \includegraphics[width=1\linewidth]{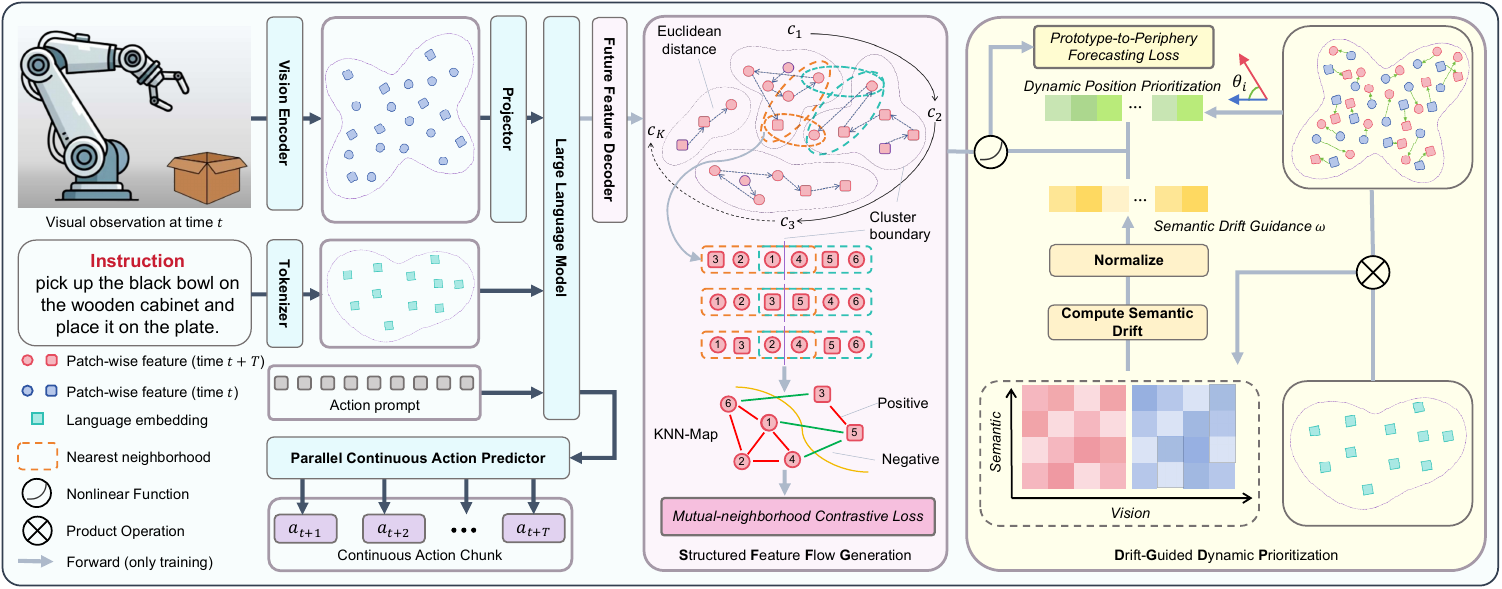}
    \vspace{-5mm}
    \caption{Overview of our LEEVLA. The \textit{Structured Feature Flow Generation} is composed of \textit{Prototype-to-Periphery (P2P)} mechanism and a \textit{Mutual-neighborhood Contrastive (MC) Loss} as shown in the pink panel. A \textit{Future Feature Decoder} predicts features at time step $t+T$ within the target latent space. We impose a \textit{P2P forecasting loss} to model structured state evolution. For each feature at $t+T$, we build a second-order $k$-NN mutual-neighborhood graph and optimize the \textit{MC loss}. The \textit{Drift-Guided Dynamic Prioritization (DGDP)} module (yellow panel) computes prioritization weight based on the \textit{Dynamic Position Prioritization (DPP)} and \textit{Semantic Drift Guidance (SDG)}. The \textit{Parallel Continuous Action Predictor (action policy)} consumes embeddings from LLM to output continuous actions.} 
    \vspace{-5mm}
    \label{fig: main}
\end{figure*}

\section{Related Works}
\subsection{Vision--Language--Action Models}
Vision--language--action (VLA) models~\cite{kim24openvla, kim2025openvla-oft, TraceVLA,zheng2025uniact,bu2025univlaRSS2025,wang2025unifiedvisionlanguageactionmodel,dreamvla25,driess2023palme, li2024cogactfoundationalvisionlanguageactionmodel} aim to map multi-view visual observations and natural-language instructions to robot actions in a single sequence modeling framework. PaLM-E \cite{driess2023palme} showed that injecting embodied signals into a pretrained language model enables grounded instruction following, and RT-2 \cite{brohan2023rt2} further demonstrated that internet-scale vision--language pretraining can transfer to real robot manipulation and arrangement tasks. More recent open efforts such as OpenVLA \cite{kim24openvla} and Octo \cite{ghosh2024octo} make this recipe accessible by standardizing the use of a VLM/LLM backbone combined with an action head over large cross-embodiment datasets, while $\pi_{0}$ \cite{black2025pio, pertsch2025fast} improve action expressivity via flow matching on top of a frozen VLM. CogACT~\cite{li2024cogactfoundationalvisionlanguageactionmodel} factorizes cognition and action by attaching a diffusion-transformer-based action module to a pretrained VLM and demonstrates excellent performance.
Recent works further improve VLA perception by selecting or compressing visual tokens. OTTER~\cite{huang2025otter} extracts instruction-aligned visual features with pretrained vision-language alignment, and Compressor-VLA~\cite{gao2025compressorvlainstructionguidedvisualtoken} compresses instruction-relevant visual tokens for efficient manipulation. In contrast, LEEVLA keeps the inference token stream unchanged and uses future-feature prediction with structured latent topology constraints during training, teaching action-conditioned environment dynamics without extra inference memory overhead.
However, existing VLA models typically treat all tokens or patches uniformly during training, which may dilute supervision on task-relevant regions~\cite{Token-Efficient}. Based on those prior works, we keep the VLA formulation. Simultaneously, we introduce drift-guided dynamic prioritization (DGDP) and a structured feature flow generation mechanism, which enable the vision-language-action model to learn where to attend and how these attended features should evolve.
\subsection{World Models}
World models~\cite{hafner2025dreamerv3, zhou2024robodreamer, hu2023gaia, bruce2024genie, wu2024ivideogpt,Li2024LWE, Learninggeneralworldmodels, Zhu2025UnifiedWM,WM_Survey} learn to forecast future states. CASCADE~\cite{Learninggeneralworldmodels} learns a world model from collaboratively gathered data across multiple agents under an information-theoretic objective motivated by Bayesian active learning. DreamerV3 \cite{hafner2025dreamerv3} showed that scalable latent dynamics can outperform specialized model-free methods across many domains. GAIA-1 \cite{hu2023gaia} performs action-conditioned world-model video generation for autonomous driving, and Genie~\cite{bruce2024genie} learns a latent action space enabling unsupervised and action-controllable interactive environments. RoboDreamer \cite{zhou2024robodreamer} brought action-conditioned video/world modeling closer to robot manipulation by factorizing objects, goals, and actions. UWM~\cite{Zhu2025UnifiedWM} integrates the action diffusion process and the video diffusion process within a unified Transformer architecture, effectively combining policy generation with robot dynamics. DreamZero~\cite{ye2026worldactionmodelszeroshot} jointly models videos and actions for zero-shot policy learning. Unlike video-generation-based control, LEEVLA uses latent-space future prediction only as auxiliary training supervision for the action policy.

\section{Preliminaries}
\label{sec:formatting}

\noindent
Following prior VLA models \cite{kim24openvla, kim2025openvla-oft}, the robot observes a third-person image $I_t^{p}\!\in\!\mathbb{R}^{H\times W\times C}$, a wrist-camera image $I_t^{w}\!\in\!\mathbb{R}^{H\times W\times C}$, proprioception $s_t$, and a language instruction $l$ at time step $t$. Here $H$, $W$ and $C$ denote the height, width, and number of channels of the visual inputs.
We adopt pretrained visual backbones (DinoV2 \cite{oquab2023dinov2} and SigLIP \cite{zhai2023siglip}) to form visual encoder $E_v(\cdot)$ that produces $N_v$ patch features $\mathbf{f}_t^{p}=E_v(I_t^{p})=\{v_{t,i}^{p}\}_{i=1}^{N_v}$ and $\mathbf{f}_t^{w}=E_v(I_t^{w})=\{v_{t,i}^{w}\}_{i=1}^{N_v}$, where each patch vector lives in $\mathbb{R}^{d_0}$ and $i$ indexes spatial locations.
A projector $\mathcal{P}:\mathbb{R}^{d_0}\!\to\!\mathbb{R}^{d}$ is applied patch-wise to obtain model-dimension tokens $\mathbf{x}_t^{p}=\mathcal{P}(\mathbf{f}_t^{p})\in\mathbb{R}^{N_v\times d}$ and $\mathbf{x}_t^{w}=\mathcal{P}(\mathbf{f}_t^{w})\in\mathbb{R}^{N_v\times d}$. 
The language instructions $l$ are encoded by a tokenizer $\Psi(\cdot)$ into $\mathbf{c}=\Psi(l)\in\mathbb{R}^{M\times d}$, and proprioception $s_t$ is embedded by $E_s(\cdot)$ into $\mathbf{x}_t^{s}=E_s(s_t)\in\mathbb{R}^{d}$. However, prior VLA models~\cite{kim24openvla, kim2025openvla-oft} struggle to handle complex tasks due to lack of perception of the future state of the environment. To incorporate reasoning of future representation, we build a future feature decoder $\mathcal{D}$, which maps the vision embedding to the future feature space: $\mathcal{D}:\mathbb{R}^{d}\!\to\!\mathbb{R}^{d_0}$. The detailed architecture of the future feature decoder is presented in the supplementary material \S 9.
The parallel continuous action predictor (action policy) generates the continuous action chunk by $\mathbf{a}_{1:T}\in\mathbb{R}^{T\times F}$, where $T$ denotes the length of the action chunk and $F$ is the action-space degrees of freedom.

\section{Methodology}

In this section, we introduce drift-guided dynamic prioritization (DGDP, \S \ref{sec: DGDP}), which consists of dynamic position prioritization (DPP) and semantic drift guidance (SDG), to automatically discover key task-relevant regions during training. Subsequently, we introduce the structured feature flow generation (SFFG, \S \ref{sec: SFFG}) strategy, allowing the model to perform structured reasoning in latent space through prototype-to-periphery (P2P) prediction, while ensuring semantic neighborhood consistency via mutual-neighborhood contrastive (MC) loss. Finally, \S \ref{sec: training_objective} details our training objective, where an $L_1$ action regression loss is combined with the P2P and MC losses to jointly optimize continuous action generation and future feature structure. We visualize an overview of our LEEVLA in Fig.~\ref{fig: main}. Detailed hyperparameter settings are included in the supplementary material. Notably, DGDP and SFFG are used only during training and do not incur any additional inference cost during test time. 

\subsection{Drift-Guided Dynamic Prioritization}
\label{sec: DGDP}

Establishing \textbf{where to attend} is essential. Paying equal attention to all information in the observation space can easily lead to gradients being diluted across static backgrounds and objects irrelevant to the task. To steer optimization direction toward task-relevant regions, we quantify feature dynamics between adjacent timestamps and estimate semantic-drift direction of each patch, then adaptively modulate its task relevance.

\begin{figure}[t]
  \centering
  \begin{subfigure}{0.49\linewidth}
    \includegraphics[width=0.95\linewidth]{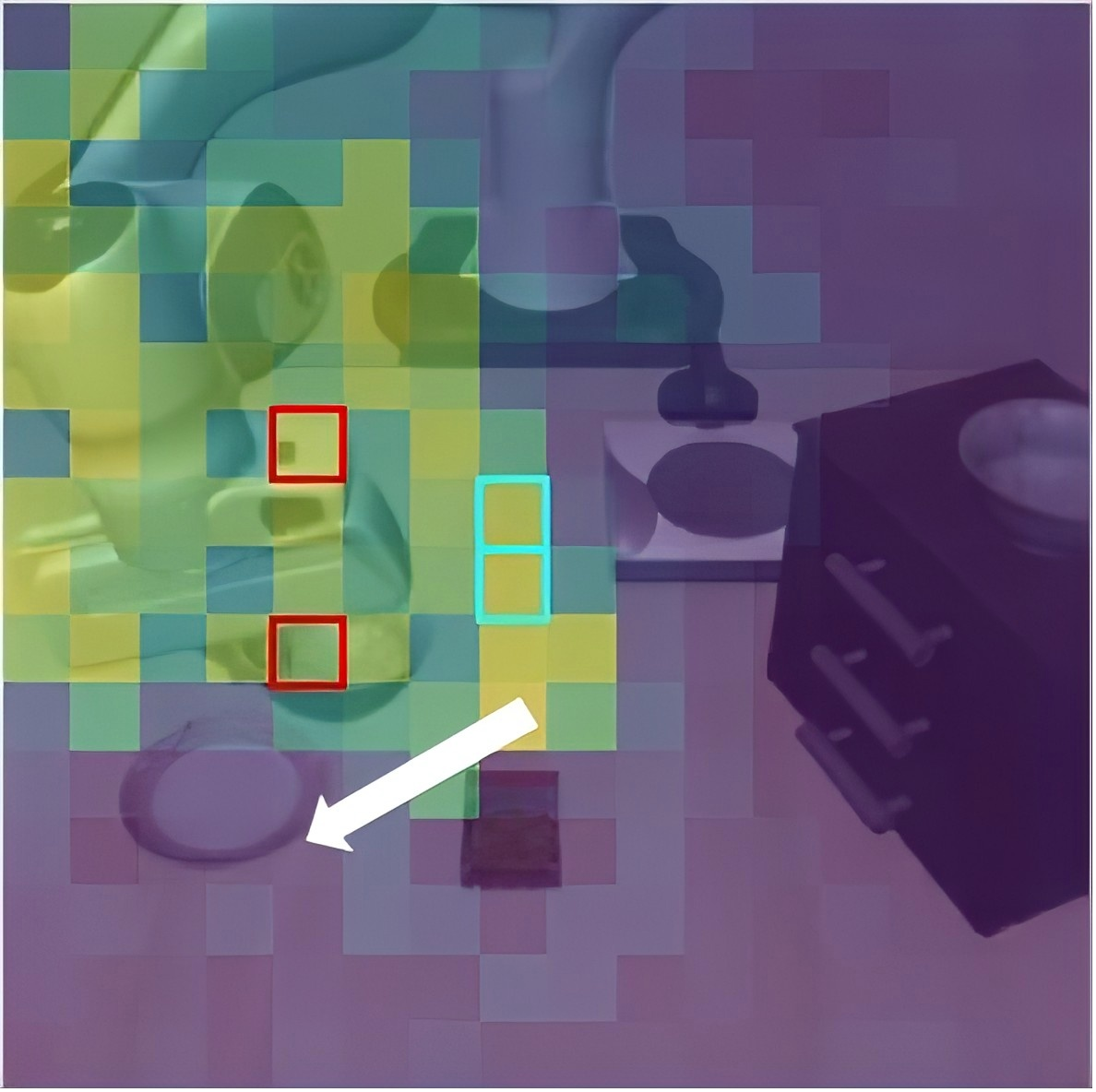}
    \caption{}
    \label{fig: DPP}
  \end{subfigure}
  \hfill
  \begin{subfigure}{0.49\linewidth}
     \includegraphics[width=0.95\linewidth]{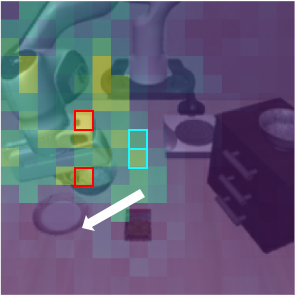}
    \caption{}
    \label{fig: DPP_SDG}
  \end{subfigure}
  \caption{Example of Weight Visualization. (a): Visualization of dynamic position prioritization (DPP). (b): Visualization of drift-guided dynamic prioritization (DGDP). The white arrow indicates the target direction of the robotic arm's movement. The semantic drift guidance factor suppresses the weights of instruction-irrelevant background regions (as indicated by the yellow box) and enhances regions with significant semantic changes along edges (as indicated by the blue box).}
  \vspace{-3mm}
  \label{fig: DGDP}
\end{figure}

\textbf{Dynamic Position Prioritization.}
When forecasting future representations in the feature space, the semantic dynamics vary markedly across spatial regions: local features associated with the robot and task-relevant objects typically exhibit strong temporal variability, whereas background or otherwise static regions remain comparatively stable. Therefore, we introduce a dynamic position prioritization mechanism that adaptively modulates prediction-loss weights at the feature level, guiding the model to focus on regions more sensitive to task execution and environment interaction. We quantify the dynamism score at each spatial position via the change in cosine similarity between adjacent time steps. For patch $i$, the feature $v_{t, i}$ at time $t$ moves to $v_{t+T, i}$ after the execution of $T$ time steps. The dynamism score $\theta_i$ is denoted as:
\begin{equation}
    \theta_i \;=\; 1-\operatorname{cos}\!\big({v}_{t,i},\,{v}_{t+T,i}\big),\quad i=1,\dots,2N_v,
\label{eq:dpp_score}
\end{equation}
where $\operatorname{cos}(\cdot)$ is the cosine similarity function. As shown in Fig. \ref{fig: DGDP} (a), the regions exhibiting salient changes in visual features between time $t$ and $t+T$ are assigned higher dynamism scores, indicating stronger attentional focus. 

\textbf{Semantic Drift Guidance.}
Focusing solely on dynamic regions ignores the directionality of task signals, leading the model to over-attend to patches whose semantics drift from task-relevant to task-irrelevant. To address this, we introduce semantic drift guidance. First, we assign an instruction-relevant score $r$ to each patch of time $t$ and $t+T$ according to the language instructions. Instruction-relevant score assignment is denoted as:
\begin{equation}
r_{t,i}\!=\!\max_{1\le j\le M}\langle\tilde{\mathbf{x}}_{t,i},\,\tilde{\mathbf{c}}_j\rangle,\;
\label{eq:pseudo_lable}
\end{equation}
where $\langle\cdot,\cdot\rangle$ denotes the Euclidean inner product, $\tilde{\mathbf{x}}_{t,i}$ is the normalized vision token of patch $i$ at time $t$, $\tilde{\mathbf{c}}_j$ denotes the normalized token of language token $j$, $M$ indicates the number of language tokens. Based on Eq. \ref{eq:pseudo_lable}, we then obtain $r_{t,i}$ and $r_{t+T,i}$  which represent the instruction-relevant scores of patch $i$ at time $t$ and $t+T$.

We define instruction-relevant semantic drift $\Delta_i$ as:
\begin{equation}
    \begin{aligned}
        \Delta_i \;=\; &\operatorname{clip} (\frac{r_{t+T,i}-r_{t,i}}{\tau},\, -\delta,\, \delta),\\
    \end{aligned}
\label{eq:drift}
\end{equation}
where $\operatorname{clip}(\cdot)$ is a numerical constraint function with a boundary of $\delta$ and $\tau$ is temperature. Both $\tau$ and $\delta$ are positive. 
To ensure a consistent dynamic range across samples and prevent domination by outliers, we normalize the instruction-relevant score to the interval [-1, 1] and map it to the semantic drift guidance factor. Semantic drift guidance factor of patch $i$ is denoted as:
\begin{equation}
\begin{aligned}
    &\omega_i \;=\; \exp (\frac{\tilde{\Delta}_i}{\tau}),&\\
\end{aligned}
\label{eq: gain}
\end{equation}
where the semantic drift $\tilde{\Delta}_i = \frac{2\cdot(\Delta_i-\min(\Delta))}{\max(\Delta)-\min(\Delta)}-1$. We can adjust the intensity of the modulation by controlling the temperature. Finally, we couple semantics and dynamics to yield the prioritization weight for each token:
\begin{equation}
\beta_i \;=\; \sigma \big(\omega_i\cdot \theta_i\big),
\label{eq:final_weight}
\end{equation}
where $\sigma(\cdot)$ is a non-linear function. In this work, we employ the sigmoid function. Fig. \ref{fig: DGDP} (b) illustrates a visualization example of DGDP weights. Through the DGDP component, the model maintains hierarchical focus over visual features, assigning higher importance to dynamic patch-level features whose semantics evolve toward the instruction-relevant, and lower importance to static features that drift toward background semantics.

\subsection{Structured Feature Flow Generation}
\label{sec: SFFG}
It is crucial to specify \textbf{how to evolve} environment representation in the latent space. Flattened token prediction, generated from top left to bottom right, corrupts the local structure of the feature space \cite{VAR, Infinity}, i.e., features from the same semantic unit are split due to sequence order, which impairs the agent's spatial reasoning ability. This disruption breaks contextual continuity. To address it, we propose structured feature flow generation (SFFG) strategy. SFFG alleviates semantic fragmentation caused by flat prediction through prototype-to-periphery (P2P) prediction mechanism. We also leverage mutual-neighborhood contrastive loss to align semantically similar features, thereby preserving the topology of the visual feature space. 

\textbf{Prototype-to-Periphery (P2P) prediction.}
We perform joint clustering on features of multi-view observations to obtain set $\mathcal{C}$, where each element represents a cluster:
\begin{equation}
\mathcal{C}\!=\!\mathcal{F}\big([\mathbf{f}_{t+T}^{p};\mathbf{f}_{t+T}^{w}])=\{c_l\}_{l=1}^{L},
\;\mu_\ell\!=\!\tfrac{1}{|c_\ell|}\!\sum_{j\in c_\ell}\mathbf{f}_{t+T,j},
\label{eq:joint_kmeans}
\end{equation}
where $\mathcal{F}(\cdot)$ represents clustering operator, $[\cdot;\cdot]$ denotes sequence concatenation along the token dimension, $L$ is the number of clusters, $\mid \cdot \mid$ indicates the set cardinality, and $\mu_\ell$ denotes centroid of cluster $c_\ell$. And then we sort members by Euclidean distance between members and centroid from nearest to farthest (prototype $\to$ periphery) within each $c_\ell$ as follows:
\begin{equation}
\overrightarrow{c_\ell}=\{v_{\varepsilon_\ell(k)}\}_{k=1}^{|c_\ell|}, \;s.t.\; \varepsilon_\ell(1)\le\dots\le\varepsilon_\ell(|c_\ell|),
\label{eq: cluster_sort}
\end{equation}
where $\varepsilon_\ell(k)=\parallel v_i - \mu_\ell\parallel_2$ denotes Euclidean distance between visual feature $v_i \in c_\ell$ and centroid $\mu_l$.

Based on the ordered sequence constructed by Eq.~\ref{eq: cluster_sort}, the P2P forecasting loss function is expressed as:
\begin{equation}
\mathcal{L}_{\text{P2P}}
\!=\! \frac{1}{|\mathcal{C}|}\!\sum_{\ell=1}^{|\mathcal{C}|}\!\frac{1}{|\overrightarrow{c_\ell}|}\!\sum_{j\in \overrightarrow{c_\ell}}^{|\overrightarrow{c_\ell}|}\!
\big(\alpha + \beta_j\big)\,\varphi \!\big(\hat{v}_{t+T,j},\,v_{t+T,j}\big),
\label{eq: obs_pred}
\end{equation}
where $\alpha$ denotes the global modulation factor to maintain attention to global information, $\varphi(\hat{v}_{t+T,j},\,v_{t+T,j})=1-\operatorname{cos}(\hat{v}_{t+T,j},\,v_{t+T,j})$ represents the cosine embedding loss, and $\hat{v}_{t+T}$ is the visual feature predicted at time $t$ for time $t+T$. In this work, we set $\alpha=1$.

\textbf{Mutual-neighborhood Contrastive (MC) Loss.}
To achieve more robust contrastive supervision under noisy clustering, we construct contrastive pairs based on high-confidence neighborhood relations in the feature space. Let $S_{ij}\!=\!\operatorname{cos}(v_{t+T,i}, v_{t+T,j})$ denote the cosine similarity between the future visual feature of samples $i$ and $j$. For each anchor $i$, we first form a first-order neighbor set: $\mathcal{G}^{(1)}_i\!=\!\operatorname{TopK}\big(\{S_{i\ell}\}_{\ell\neq i}\big)$ which keeps the $K$ most similar tokens to $i$. To further enlarge the pool of potentially clean positives while still staying in a locally consistent region, we define a second-order neighbor set: $\mathcal{G}^{(2)}_i\!=\!\bigcup_{j\in\mathcal{G}^{(1)}_i}\operatorname{TopM}\big(\{S_{jr}\}_{r\neq j}\big)$, i.e. the union of the 
$M$ nearest neighbors of each first-order neighbor. We set $K=10$ and $M=5$. We then select only those tokens that are mutual neighbors to suppress the asymmetric or spurious links introduced by clustering noise. Concretely, the positive set for anchor $i$ is:
\begin{equation}
    \mathcal{G}_i^+=\{\,j\mid j\in\mathcal{G}^{(1)}_i\cup\mathcal{G}^{(2)}_i,\; i\in\mathcal{G}^{(1)}_j\cup\mathcal{G}^{(2)}_j\,\}.
    \label{eq: pos_set}
\end{equation}

We adopt the InfoNCE loss~\cite{oord2018infonce, chen2020simple} over these mutual-neighborhood positives to pull them closer in the feature space. 
The mutual-neighborhood contrastive loss is calculated as a function of the similarity relationships among samples within their respective neighborhoods in the representation space, and is formally defined as follows:
\begin{equation}
\mathcal{L}_{\text{MC}}
\!=\! -\! \frac{1}{|\mathcal{I}|}\!\sum_{i\in\mathcal{I}}\!
\frac{1}{|\mathcal{G}_i^+|}\!\sum_{j\in\mathcal{G}_i^+}
\!\log\!\frac{\exp\!\big(S_{ij}/\tau_c\big)}{\!\sum_{\ell\ \! \in \mathcal{V}\setminus\!\{i\}}\!\exp\!\big(S_{i\ell}/\tau_c\big)},\!
\label{eq:contrastive_loss}
\end{equation}
where $\mathcal{I}\!=\!\{i\!\in\!\mathcal{V}\!\mid \!|\mathcal{G}_i^+|\!>\!0\}$ represents non-empty set of positive samples, $\mathcal{V}\!=\! \{1,\!\dots\!,\! 2N_v\}$ is index set representing multi-view visual features, and $\tau_c$ is the temperature. Rather than simply enlarging the neighborhood size, the mutual-neighborhood mechanism adaptively identifies high-confidence and symmetric feature relations, providing more stable supervision under noisy and weakly labeled robot demonstration data.

\subsection{Training Objective}
\label{sec: training_objective}
Similar to \cite{kim2025openvla-oft}, we adopt an $L_1$ regression strategy and employ parallel action decoding, which is efficient and tends to produce more accurate actions. The action policy is an MLP head that directly regresses continuous actions from the last-layer hidden states of the large language model. Training minimizes the average $L_1$ distance to ground truth actions to filter noise from the training demonstrations~\cite{kim2025openvla-oft}. The action prediction loss is calculated as :
\begin{equation}
\mathcal{L}_{\text{action}}
\!=\! \frac{1}{T}\sum_{i=1}^{T}  \left| a_{t+i} - \hat{a}_{t+i} \right|_1,\!
\label{eq:l1_loss}
\end{equation}
where $\hat{a}_{t+i}$ represents the predicted action of time $t+i$ and $\left|\cdot\right|$ is the $L_1$ norm.
The overall training objective is expressed as: 
\begin{equation}
    \mathcal{L}_{\text{total}} = \lambda_1 \mathcal{L}_{\text{action}} +  \lambda_2 \mathcal{L}_{\text{P2P}} +  \lambda_3 \mathcal{L}_{\text{MC}},
    \label{eq:total_loss}
\end{equation}
where  $\lambda_1$,  $\lambda_2$, and  $\lambda_3$ are hyperparameters that balance the contributions of the action regression loss, prototype-to-periphery (P2P) forecasting loss, and mutual-neighborhood contrastive (MC) loss. 

\begin{table*}[t]
\centering
\renewcommand{\arraystretch}{1}
\setlength{\tabcolsep}{1.6mm}
\footnotesize
\begin{NiceTabular}{
@{}ccc|cccccc@{}
}
\toprule
\multicolumn{1}{c}{\multirow{2}{*}{Methods}} &
\multicolumn{1}{c}{\multirow{2}{*}{Reasoning}} &
\multicolumn{1}{c|}{\multirow{2}{*}{Size}} &
\multicolumn{6}{c}{CALVIN Results} \\
\multicolumn{1}{c}{} & \multicolumn{1}{c}{} & \multicolumn{1}{c|}{} &
\multicolumn{1}{c}{1} & \multicolumn{1}{c}{2} &
\multicolumn{1}{c}{3} & \multicolumn{1}{c}{4} &
\multicolumn{1}{c}{5} & \multicolumn{1}{c}{Avg.(\%)} \\ \midrule
OpenVLA (CoRL'2025) \cite{kim24openvla} & \ding{55} & 7B & 91.3 & 77.8 & 62.0 & 52.1 & 43.5 & 3.80 \\
$\pi_0$ (RSS'2025) \cite{black2025pio} & \ding{55} & 3B & 94.3 & 87.0  & 77.9 & 68.5 & 59.4 & 3.87 \\
$\pi_{0.5}$ (RSS'2025) \cite{black2025pio} & \ding{55} & 3B & 91.9 & 84.6 & 79.4 & 75.5 & 71.0 & 4.02 \\
OpenVLA-OFT (RSS'2025) \cite{kim2025openvla-oft} & \ding{55} & 7B & 96.3 & 89.1 & 82.4 & 75.8 & 66.5 & 4.10 \\
UniVLA (RSS'2025) \cite{bu2025univlaRSS2025} & \ding{55} & 7B & 95.5 & 85.5 & 75.4 & 66.9 & 56.5 & 3.80 \\
\midrule
\rowcolor{mygray}
LEEVLA-large (Ours) & \ding{51} & 7B &
\bf\color{deepred}98.8 & \bf\color{deepred}94.5 & \bf\color{deepred}87.3 & \bf\color{deepred}80.6 & \bf\color{deepred}72.7 & \bf\color{deepred}4.34 \\
\bottomrule
\end{NiceTabular}
\caption{\textbf{The success rates of large-scale models on the CALVIN benchmark.}
We evaluate LEEVLA-large on four CALVIN ABC-D tasks and report the success rate for each task and the average length.
}
\label{tab: exp_calvin}
\end{table*}

\begin{table*}[t]
\centering
\renewcommand{\arraystretch}{1}
\setlength{\tabcolsep}{1.6mm}
\footnotesize

\begin{NiceTabular}{
@{\extracolsep{\fill}}c l c c | c c c c c@{}
}
\toprule
\multirow{2}{*}{Scale} &
\multirow{2}{*}{Method} &
\multirow{2}{*}{Reasoning} &
\multirow{2}{*}{Size} &
\multicolumn{5}{c}{LIBERO Results} \\
& & & &
Spatial(\%) & Object(\%) & Goal(\%) & Long(\%) & Average(\%) \\
\midrule

\multirow{6}{*}{Small}
& Octo (RSS'2024) \cite{ghosh2024octo}
& \ding{55} & 0.1B & 78.9 & 85.7 & 84.6 & 51.1 & 75.1 \\

& UniACT (CVPR'2025) \cite{zheng2025uniact}
& \ding{55} & 0.5B & 77.0 & 87.0 & 77.0 & 70.0 & 77.8 \\

& Seer (ICLR'2025) \cite{seer}
& \ding{55} & 0.3B & -- & -- & -- & 87.7 & -- \\

& DreamVLA (NeurIPS'2025) \cite{dreamvla25}
& \ding{51} & 0.3B & 97.5 & 94.0 & 89.5 & 89.5 & 92.6 \\

& FLOWER (CoRL'2025) \cite{reuss2025flower}
& \ding{55} & 1B & 97.1 & 96.7 & 95.6 & 93.5 & 95.7 \\

\rowcolor{mygray}
& LEEVLA-mini (Ours)
& \ding{51} & 0.5B
& \bf\color{deepred}98.6
& \bf\color{deepred}99.0
& \bf\color{deepred}97.0
& \bf\color{deepred}95.5
& \bf\color{deepred}97.5 \\

\midrule

\multirow{7}{*}{Large}
& OpenVLA (CoRL'2024) \cite{kim24openvla}
& \ding{55} & 7B & 84.7 & 88.4 & 79.2 & 53.7 & 76.5 \\

& CoT-VLA (CVPR'2025) \cite{zhao2025cotvla}
& \ding{51} & 7B & 87.5 & 91.6 & 87.6 & 69.0 & 81.1 \\

& $\pi_0$ (RSS'2025) \cite{black2025pio}
& \ding{55} & 3B & 96.8 & 98.8 & 95.8 & 85.2 & 94.1 \\

& $\pi_{0.5}$ (CoRL'2025) \cite{black2025pio}
& \ding{55} & 3B & 97.0 & 99.0 & 98.0 & 96.0 & 97.5 \\

& OpenVLA-OFT (RSS'2025) \cite{kim2025openvla-oft}
& \ding{55} & 7B & 97.6 & 98.4 & 97.9 & 94.5 & 97.1 \\

& UniVLA (RSS'2025) \cite{bu2025univlaRSS2025}
& \ding{55} & 7B & 96.5 & 96.8 & 95.6 & 92.0 & 95.2 \\

\rowcolor{mygray}
& LEEVLA-large (Ours)
& \ding{51} & 7B
& \bf\color{deepred}98.8
& \bf\color{deepred}99.0
& \bf\color{deepred}98.6
& \bf\color{deepred}96.4
& \bf\color{deepred}98.2 \\

\bottomrule
\end{NiceTabular}

\caption{\textbf{Success rates on the LIBERO benchmark.}
We evaluate LEEVLA-mini and LEEVLA-large on four LIBERO tasks and report the success rate for each task and the average success rate.
\emph{Reasoning} indicates whether the model performs an explicit reasoning stage before action generation, and \emph{Size} denotes the parameter scale of the language backbone.}
\vspace{-2mm}
\label{tab:exp_libero}
\end{table*}

\section{Experiment}

\subsection{Implementation Details}

LEEVLA-large is initialized from OpenVLA-7B and further pretrained on a large mixture of datasets from Open X-Embodiment~\cite{openxembodiment2024}, which covers diverse robot and vision--language trajectories. 
We train LEEVLA-large for 50k to 150k optimization steps, where more challenging tasks typically require longer training schedules.
LEEVLA-mini is initialized from miniVLA~\cite{kim24openvla, belkhale2024minivla}, which is pretrained on LIBERO-90~\cite{liu2023liberobenchmarkingknowledgetransfer}, and we train LEEVLA-mini for 20k to 50k steps. For LEEVLA-large, we use a learning rate of $5\times10^{-4}$; for LEEVLA-mini, we use $2\times10^{-5}$. All models are optimized using an AdamW optimizer~\cite{Loshchilov2017AdamW}, with both training and inference performed on a computing infrastructure equipped with \(\,8\times\) A100 (80\,GB) GPUs.
Detailed hyperparameters are provided in the supplementary materials \S 8. 
For real-world experiments, we adopt Universal Robots UR5 collaborative robotic arm, which has 6 degrees of freedom. The experiments require the robot to complete three tasks: placing an object, pressing a button, and closing a drawer. Each experimental setup is evaluated over 20 consecutive trials.

\subsection{Benchmark}

We compare our method against representative VLA systems on the LIBERO suite \cite{liu2023liberobenchmarkingknowledgetransfer}, which groups manipulation tasks into four categories: \emph{Spatial}, \emph{Object}, \emph{Goal}, and \emph{Long}. We report success rates of each task and the average success rate.
We report the success rate for each task and the overall average success rate across the 10 language instructions and 50 episodes under 3 random seeds.

\textbf{Small-scale baselines.} Tab.~\ref{tab:exp_libero} compares our LEEVLA-mini with recent small-scale VLAs whose sizes are less than 1 billion parameters. Octo \cite{ghosh2024octo} is an open-source generalist policy for robotic manipulation which pretrained on the Open X-Embodiment trajectories. UniACT \cite{zheng2025uniact} builds an embodied foundation model in a universal action space. Seer \cite{seer} is an end-to-end Predictive Inverse Dynamics Model that jointly performs conditional visual foresight and inverse-dynamics action prediction. DreamVLA \cite{dreamvla25} introduces explicit reasoning by forecasting visual goals before action decoding. FLOWER \cite{reuss2025flower} is a 950M-parameter VLA policy that improves the efficiency of action generation. Our LEEVLA-mini includes an explicit reasoning stage through structured feature flow generation, which is reflected in the \emph{Reasoning} column. This set isolates the effect of reasoning and token prioritization at similar parameter budgets.

\textbf{Large-scale baselines.} Tab.~\ref{tab:exp_libero} shows the comparison results between LEEVLA-large and prior large-scale models. OpenVLA~\cite{kim24openvla} is a widely used 7B open-source baseline built on Llama-2-7B~\cite{touvron2023llama} with DINOv2~\cite{oquab2023dinov2} and SigLIP~\cite{zhai2023siglip} vision features. $\pi_{0}$ \cite{black2025pio} adopts a pretrained VLM (PaliGemma) with a flow-matching action expert and action chunking for continuous control. OpenVLA-OFT \cite{kim2025openvla-oft} instantiates an Optimized Fine-Tuning recipe for OpenVLA~\cite{kim24openvla} with parallel decoding, chunked continuous actions, and an $L_1$ regressive policy. 
UniVLA~\cite{bu2025univlaRSS2025} learns cross-embodiment VLA policies by extracting task-centric latent action representations from large-scale, heterogeneous videos and decoding them into robot-specific actions.
MemoryVLA \cite{shi2025memoryvlaperceptualcognitivememoryvisionlanguageaction} adds a perceptual-cognitive memory to handle long-horizon temporal dependence. We mark \emph{Reasoning} according to whether a method introduces an explicit intermediate stage before action output.

\begin{figure}[h]
  \centering
    \includegraphics[width=1\linewidth]{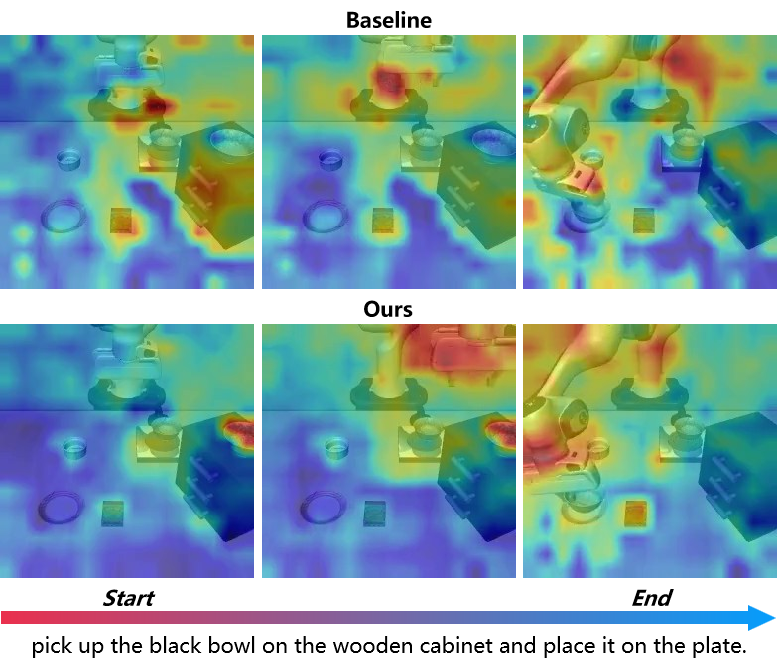}
  \caption{Comparison of instruction-relevance between LEEVLA-mini and baseline vision features. We present a visualization of the cosine similarity between vision embeddings and language instruction embeddings. ``Baseline" denotes the miniaturized model variant where both DGDP and SFFG components are ablated.}
  \vspace{-5mm}
  \label{fig: comparation_vis}
\end{figure}

\subsection{Experimental Results}
\textbf{Simulation Environment Results.} As shown in Tab.~\ref{tab:exp_libero}, LEEVLA effectively adapts to various task settings of LIBERO, achieving optimal or competitive performance across most task suites. 

In Fig.~\ref{fig: comparation_vis}, we further visualize the correlation between vision features and the instruction by computing the cosine similarity between the visual and instruction embeddings. The top part of Fig.~\ref{fig: comparation_vis} (Baseline) indicates that a model that does not infer future states of the environment fails to leverage visual embeddings to effectively guide action generation. 
Benefiting from the SFFG and DGDP modules, LEEVLA achieves a much tighter coupling between visual observations and action generation.

\begin{figure}
  \centering
    \includegraphics[width=1\linewidth]{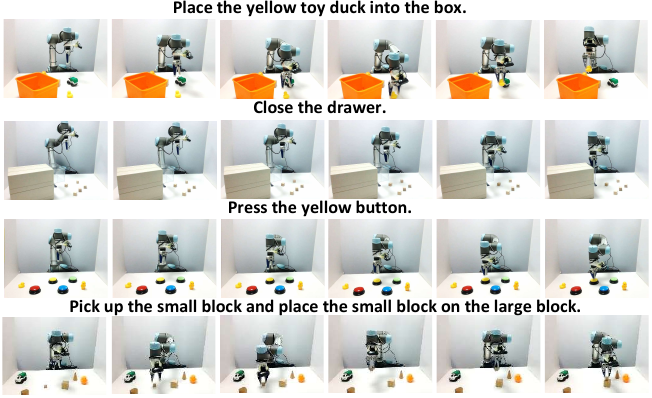}
  \caption{The qualitative results of LEEVLA in real-world environment. We set up three real-world tasks to demonstrate the generalization performance of our model in real world. 
  }
\label{fig: real_world}
\vspace{-2mm}
\end{figure}

\textbf{Real-world Results.} We provide a quantitative analysis in real-world settings, as shown in Table~\ref{tab: real-world}. Across multiple tasks, our approach consistently outperforms OpenVLA. Additionally, we offer qualitative insights through visualizations, as illustrated in Fig.~\ref{fig: real_world}. 

\begin{table}[t]
\centering
\renewcommand{\arraystretch}{1}
\setlength{\tabcolsep}{1.2mm}
\footnotesize
\begin{NiceTabular}{@{}l|ccccc@{}}
\toprule
Methods & Place& Press& Drawer&Long& Average\\ \midrule
OpenVLA~\cite{kim24openvla}&       30&               55&        40 & 35 &            40\\
LEEVLA-large (Ours)&       70&               80&        65& 60 &             78.5\\ \bottomrule
\end{NiceTabular}
\vspace{-3mm}
\caption{Real-world performance comparison on manipulation tasks. We evaluate LEEVLA-large (Ours) and OpenVLA in real-world environments across three representative tasks: \textit{Place}, \textit{Press}, \textit{Drawer}, and their overall average success rate (\%).}
\label{tab: real-world}
\vspace{-5mm}
\end{table}

\begin{table}[!t]
\centering
\renewcommand{\arraystretch}{1.05}
\setlength{\tabcolsep}{1.2mm}
\footnotesize
\begin{NiceTabular}{l|ccc}
\toprule[0.6pt]
Stage & Component & Peak Memory (GB) & Latency (ms) \\
\midrule

\multirow{4}{*}{Train}
& P2P & 0.237 & 2.01 \\
& MC & 0.652 & 1200.91 \\
& DPP & 0.726 & 9.60 \\
& SDG & 1.052 & 22.68 \\
\midrule
\multirow{3}{*}{Inference}
& OpenVLA-OFT & 15.639 & 124.03 \\
& LEEVLA-large & 15.639 & 124.36 \\
& LEEVLA-mini & 5.088 & 48.61 \\
\bottomrule[0.6pt]
\end{NiceTabular}
\vspace{-3mm}
\caption{Complexity analysis of different components during training and inference. We report the peak memory usage and latency of P2P, MC, DPP, and SDG.}
\label{tab:complexity_analysis}
\vspace{-5mm}
\end{table}

\subsection{Ablation study}
In this section, we conduct a series of ablations on LIBERO using LEEVLA-mini to better understand the contribution of each component in LEEVLA. As shown in Tab.~\ref{tab:ablation}, each proposed component brings a consistent improvement over the baseline. The base LEEVLA-mini model without prototype-to-periphery (P2P), mutual-neighborhood contrastive (MC) loss, dynamic position prioritization (DPP), or semantic-drift guidance (SDG) achieves a success rate of 94.8\%. Introducing P2P prediction alone improves performance to 95.2\% (+0.4), indicating that enforcing an ordered feature flow is beneficial for policy learning. Adding MC loss further boosts the success rate to 95.6\% (+0.8 over baseline), suggesting that preserving local semantic topology in latent space stabilizes future feature prediction. On top of this structured feature flow generation, enabling DPP yields the largest single gain, reaching 96.3\% (+1.5 over baseline), which highlights the importance of concentrating supervision on interaction-centric regions. Finally, incorporating the SDG leads to the best performance of 96.6\%. The experiments demonstrate that SFFG (P2P+MC) and DGDP (DPP+SDG) are complementary, jointly contributing to more accurate and robust action policies.

\begin{table*}[!t]
\centering
\renewcommand{\arraystretch}{0.9}
\setlength{\tabcolsep}{2.5mm}
\tiny
\resizebox{\textwidth}{!}{
\begin{NiceTabular}{@{}cc|cc|c|*{10}{c}@{}}
\toprule[0.6pt]
\multicolumn{2}{c|}{SFFG} 
& \multicolumn{2}{c|}{DGDP} 
& \multirow{2}{*}{Avg. Suc. (\%)} 
& \multicolumn{10}{c}{Instructions} \\
P2P & MC & DPP & SDG 
& 
& I1 & I2 & I3 & I4 & I5 & I6 & I7 & I8 & I9 & I10 \\ 
\midrule
\textcolor{lightgray}{\ding{55}} 
& \textcolor{lightgray}{\ding{55}}
& \textcolor{lightgray}{\ding{55}}
& \textcolor{lightgray}{\ding{55}}
& 94.8
& 90.0 & 96.0 & 98.0 & 90.0 & 96.0 & 98.0 & 80.0 & 100.0 & 100.0 & 100.0 \\

\ding{51}
& \textcolor{lightgray}{\ding{55}}
& \textcolor{lightgray}{\ding{55}}
& \textcolor{lightgray}{\ding{55}}
& 95.2
& 96.0 & 100.0 & 92.0 & 92.0 & 100.0 & 100.0 & 76.0 & 98.0 & 98.0 & 100.0 \\

\ding{51}
& \ding{51}
& \textcolor{lightgray}{\ding{55}}
& \textcolor{lightgray}{\ding{55}}
& 95.6
& 96.0 & 94.0 & 94.0 & 92.0 & 100.0 & 100.0 & 84.0 & 100.0 & 98.0 & 98.0 \\

\ding{51}
& \ding{51}
& \ding{51}
& \textcolor{lightgray}{\ding{55}}
& 96.3
& 98.0 & 98.0 & 94.7 & 87.3 & 100.0 & 97.3 & 89.3 & 100.0 & 100.0 & 98.7 \\

\rowcolor{mygray}
\ding{51}
& \ding{51}
& \ding{51}
& \ding{51}
& 97.0
& 100.0 & 100.0 & 98.0 & 88.0 & 100.0 & 100.0 & 94.0 & 100.0 & 98.0 & 92.0 \\
\bottomrule[0.6pt]
\end{NiceTabular}
}
\vspace{-3mm}
\caption{Results of the ablation study on LIBERO-Goal. We investigate the contribution of each component by progressively ablating the four components: P2P, MC, DPP, and SDG.}
\label{tab:ablation}
\vspace{-5mm}
\end{table*}

\begin{table}[!t]
\centering
\renewcommand{\arraystretch}{1}
\setlength{\tabcolsep}{1.2mm}
\footnotesize
\begin{NiceTabular}{c|c}
\toprule
Method & Suc. (\%) \\ 
\midrule
SFFG w/o reorder & 94.7 \\
\rowcolor{mygray}
SFFG w/ reorder & 95.2 \\
\bottomrule
\end{NiceTabular}
\vspace{-3mm}
\caption{Effect of feature reordering on future feature prediction. We evaluate LEEVLA-mini on LIBERO-Goal with and without the feature reordering module, using only the ``Baseline+P2P"}
\vspace{-5mm}
\label{tab:reorder_ablation}
\end{table}

\begin{table}[!t]
\centering
\renewcommand{\arraystretch}{1}
\setlength{\tabcolsep}{1.2mm}
\footnotesize
\begin{NiceTabular}{c|c}
\toprule
Method & Suc. (\%) \\ 
\midrule
LEEVLA-mini w/o $\alpha$ & 95.8 \\
\rowcolor{mygray}
LEEVLA-mini w/ $\alpha$ & 97.0 \\
\bottomrule
\end{NiceTabular}
\vspace{-3mm}
\caption{Performance comparison between using the global factor $\alpha$ and not using the global factor $\alpha$. We tested the effect of the global modulation factor $\alpha$ on the action policy.}
\vspace{-5mm}
\label{tab:alpha_ablation}
\end{table}

\begin{figure}[!ht]
  \centering
  
  \begin{subfigure}{0.49\linewidth}
    \includegraphics[width=1\linewidth]{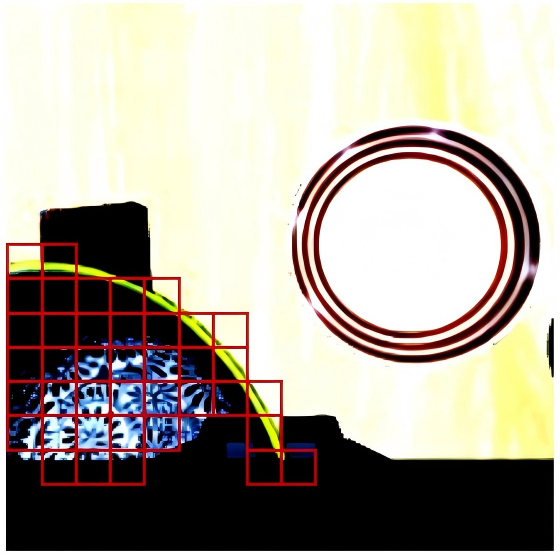}
    \caption{}
    \label{fig: bowl}
  \end{subfigure}
  \hfill
  \begin{subfigure}{0.49\linewidth}
     \includegraphics[width=1\linewidth]{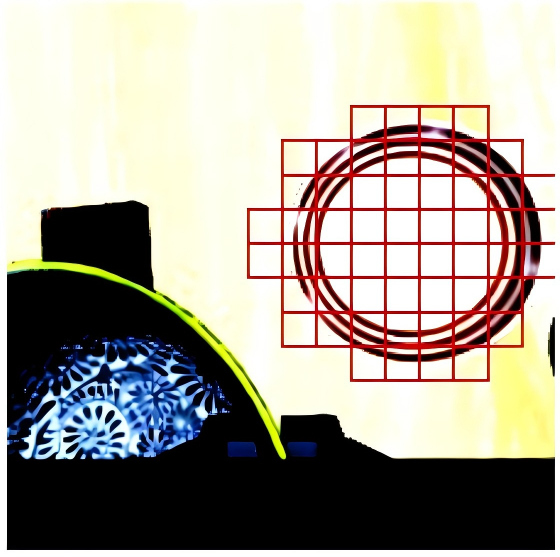}
    \caption{}
    \label{fig: plate}
  \end{subfigure}
  \hfill
  \begin{subfigure}{0.49\linewidth}
     \includegraphics[width=1\linewidth]{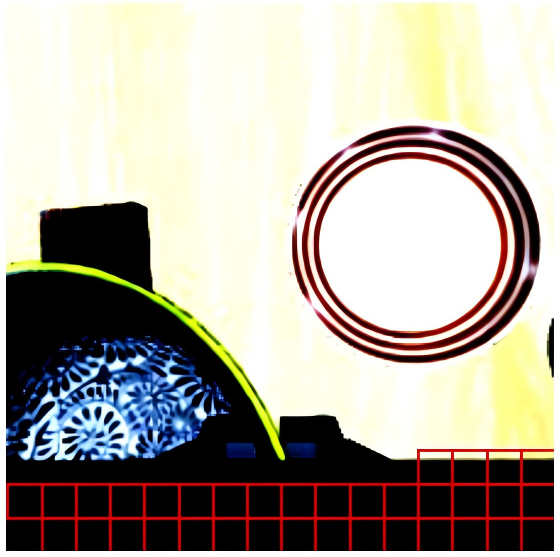}
    \caption{}
    \label{fig: robot}
  \end{subfigure}
  \hfill
  \begin{subfigure}{0.49\linewidth}
     \includegraphics[width=1\linewidth]{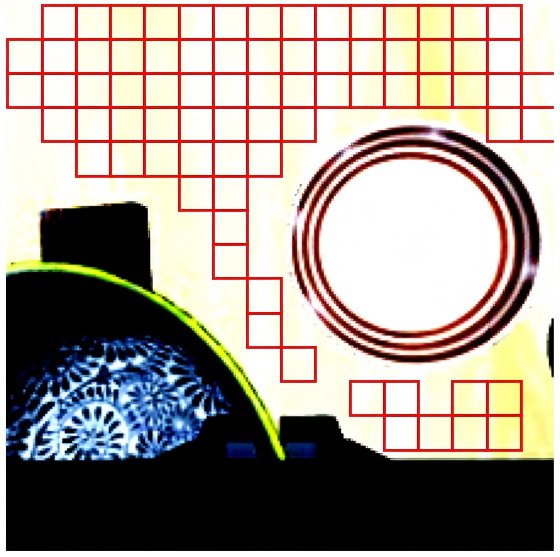}
    \caption{}
    \label{fig: table}
  \end{subfigure}
  \vspace{-3mm}
  \caption{Wrist-View Image Clustering Visualization Results. We present the actual patches corresponding to different clusters after clustering, where (a), (b), (c), and (d) represent a bowl, a plate, a gripper, and a tabletop, respectively. It can be observed that within the set of encoded image patch features, patches with similar semantics are grouped into the same feature cluster.}
  \vspace{-15mm}
  \label{fig: cluster_vis}\
\end{figure}

\section{Discussion}
\textbf{Why do we encourage models to reason in the structured latent space?} 

Human-selected external conditions capture a narrow and specific concept of the environment, often misaligned with the clues the model actually uses. Pre-trained visual features capture more structural signals. Our SFFG imposes a prototype-to-periphery ordering so tokens from the same semantic unit are predicted together, preserving spatial semantic continuity and improving long-horizon prediction. As shown in Fig. \ref{fig: cluster_vis}, LEEVLA considers the structured information between different patches.

\textbf{Why do we need to reorder the visual features?} 
Flat token prediction processes visual tokens in a fixed order, which ignores how features are actually organized in the latent space. As a result, the model reasons within a discontinuous semantic space, compromising generalization. As shown in \S\ref{sec: SFFG}, tokens that belong to the same semantic unit can be far apart in the flattened sequence, even though they are close in feature space. This mismatch breaks local contextual continuity and makes it harder for the policy to reason about spatially coherent changes. Tab.~\ref{tab:reorder_ablation} shows the effect of feature reordering on future feature prediction.

\textbf{Why do we use the global factor $\alpha$ instead of relying only on prioritization weights $\beta$?} 
Intuitively, $\beta$ amplifies task-relevant tokens. Without the global factor $\alpha$ and relying solely on $\beta$, the model becomes overly selective: contact regions are over-emphasized, while background tokens are almost discarded. However, background in manipulation scenes provides crucial spatial context (e.g., table boundaries, obstacles, robot base) that is important for geometry and long-horizon feasibility. The global factor $\alpha$ ensures that even down-weighted regions retain a weak but non-zero contribution, preserving global layout. As shown in Tab.~\ref{tab:alpha_ablation}, using both $\alpha$ and $\beta$ enables the model to focus on task-critical areas without losing overall scene awareness, whereas the $\beta$-only variant tends to over-focus and degrades performance. Therefore, we train LEEVLA with both $\alpha$ and $\beta$ learn to sharply highlight task-critical regions while still maintaining understanding of the whole environment.

\section{Conclusion}

We introduce LEEVLA for reasoning in latent feature space. 
By forecasting structured future features, LEEVLA exploits the relational structure already encoded in the visual backbone and avoids hand-crafted hypothesis spaces. Our structured feature flow generation (SFFG) treats prediction as a latent state transition: prototype-to-periphery (P2P) anchors the flow on robust prototypes before refining toward cluster periphery, while mutual-neighborhood contrastive (MC) loss preserves local topology by emphasizing reciprocal neighbors. Complementing this, drift-guided dynamic prioritization (DGDP) component of dynamic position prioritization (DPP) and semantic drift guidance (SDG) focuses supervision on dynamically active, instruction-relevant patches, reducing the impact of static background. Evaluated at two scales, LEEVLA-mini (0.5B) and LEEVLA-large (7B) achieve the state-of-the-art performance on the LIBERO and Calvin benchmark.


{
    \small
    \bibliographystyle{ieeenat_fullname}
    \bibliography{main}
}


\end{document}